\documentclass[a4paper, 10pt, conference]{ieeeconf}      

\IEEEoverridecommandlockouts                              
                                                          
\usepackage{url}
\usepackage{comment}
\usepackage{amsmath,amssymb,amsfonts, mathtools}
\usepackage{graphicx}
\usepackage{bm}
\usepackage{epstopdf}
\usepackage{placeins}
\usepackage{bm}
\usepackage{cite}
\let\labelindent\relax
\usepackage[inline]{enumitem}
\usepackage{multicol}
\usepackage{tikz}
\usepackage{caption}
\usepackage{subcaption}
\usepackage{tikz}
\usepackage{svg}
\usepackage[linesnumbered,ruled,vlined]{algorithm2e}
\SetKwInput{KwInput}{Input}    
\SetKwInput{KwOutput}{Output}  

\newcommand{\vect}[1]{\boldsymbol{#1}}

\DeclareMathOperator*{\argmin}{arg\,min}

\title{\LARGE \bf A Framework for Learning and Reusing Robotic Skills}

\author{Brendan Hertel, Nhu Tran, Meriem Elkoudi and Reza Azadeh
	\thanks{Authors are with the Persistent Autonomy and Robot Learning (PeARL) Lab, University of Massachusetts Lowell, Lowell, MA 01854, USA. Emails: \tt{\{brendan\_hertel, nhu\_tran, meriem\_elkoudi\}@student.uml.edu, reza@cs.uml.edu}}
 }
 
\begin{document}

\maketitle
\thispagestyle{empty}
\pagestyle{empty}

\begin{abstract}
    In this paper, we present our work in progress towards creating a library of motion primitives. This library facilitates easier and more intuitive learning and reusing of robotic skills. Users can teach robots complex skills through Learning from Demonstration, which is automatically segmented into primitives and stored in clusters of similar skills. We propose a novel multimodal segmentation method as well as a novel trajectory clustering method. Then, when needed for reuse, we transform primitives into new environments using trajectory editing. We present simulated results for our framework with demonstrations taken on real-world robots.
\end{abstract}

\section{Introduction}
\label{sec:intro}

As robots enter into the daily lives of users all around the world, the ability for robots to learn novel skills must be intuitive. One step towards achieving this goal is implementing a library of motion primitives~\cite{pastorDMP2009}. This library would be a repository of demonstrated primitives such as reaching, placing, or pushing. When robots perform a complex task, it can be executed by stringing together a series of motion primitives. However, teaching robots these motion primitives in every context is time-consuming and difficult for users. If robots could be taught primitives for one skill in one environment, construct a skill model, and transfer this skill model across environments and contexts, it would be more intuitive and efficient when teaching robots new skills.

There are three essential tasks to a library of motion primitives: learning, encoding, and reusing skills. To learn skills, we assume skills are taught through Learning from Demonstration (LfD)~\cite{Argall2009survey}. If a primitive skill is demonstrated, it can be immediately encoded. However, demonstrations may be more complex and are required to be broken down into motion primitives. To do this, we apply segmentation to the captured demonstrations. However, we consider multiple modes of interaction in our segmentation, and combine segmented modes probabilistically. In order to differentiate demonstrations, we cluster similar primitives together. Similar skills are put in the same cluster, such that if a reproduction of that skill is required, one or more demonstrations can be recalled. We propose a novel trajectory clustering algorithm that can automatically find the number of clusters, whereas many other clustering algorithms require this as a parameter. Finally, to reuse skills, we employ a trajectory editing method~\cite{Nierhoff2016LTE} to transform skills into novel environments. This method allows for the shape of primitives to be maintained across different contexts.

\section{Methodology} 
\label{sec:method}

\begin{figure}[t]
    \centering
    \includegraphics[width=0.98\columnwidth]{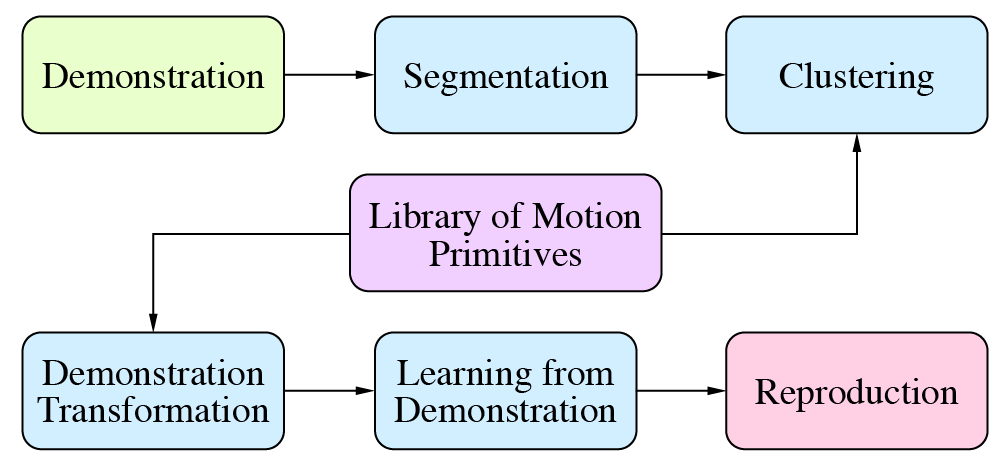}
    \caption{\small{The proposed framework for creating a library of motion primitives.}}
    \label{fig:library-flow}
\end{figure}

In this paper, we propose a framework for creating a library of reusable motion primitives, shown in Fig.~\ref{fig:library-flow}. There are several challenges in creating a motion primitive library including
\begin{enumerate*}[label=(\roman*)]
  \item discovery of motion primitives from complex demonstrations
  \item clustering found motion primitives, and
  \item selection of one or more demonstrated primitives from the motion library for use in Learning from Demonstration.
\end{enumerate*}
To address these challenges, we use several modules in the motion primitive library. First, for the discovery of motion primitives, we use multimodal segmentation of demonstrations. Whereas many previous methods~\cite{nakamura2017segmenting} segment only one source of data, we probabilistically combine data from joint space, task space, and sensors to create a segmentation method for breaking down complex tasks into motion primitives. Then, we propose a novel trajectory clustering method based on elastic maps~\cite{gorban_zinovyev}, which uses the energy associated with the elastic maps to automatically determine the number of clusters. In this application, the number of clusters is equivalent to the number of distinct motion primitives known in the motion primitive library. Finally, we use a context-based selection method for selecting reusable motion primitives from the library. Once the known primitives are selected, they are provided with a Learning from Demonstration representation to create a reproduction, which is then executed by the robot. The following sections will provide further details on the novel aspects of our framework.

\subsection{Segmentation}

\begin{figure*}[ht]
    \centering
    \includegraphics[width=0.9\linewidth]{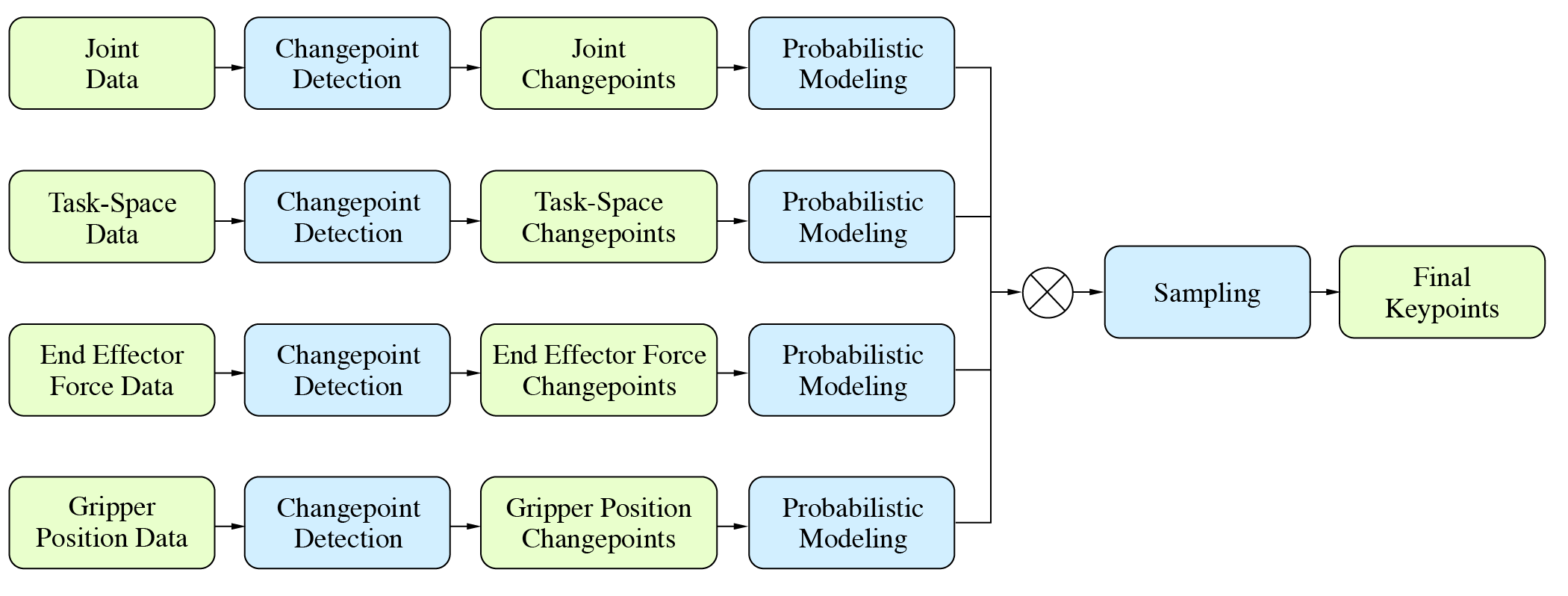}
    \caption{\small{The proposed probabilistic multimodal segmentation method.}}
    \label{fig:segmentation-flow}
\end{figure*}

Our first contribution is the use of probabilistic multimodal segmentation\footnote{Segmentation module available at: \url{https://github.com/brenhertel/Probabilistic-Segmentation/}} to break down complex tasks into one or more motion primitives, shown in Fig.~\ref{fig:segmentation-flow}. Given a single demonstration $\mathcal{D}$, often including multiple data streams such as joint data, task-space data, etc., we segment each of these individual data streams $d$ separately, then probabilistically combine the keypoints $k$ found from each of these segmentations. For the segmentation of an individual data stream, we construct segments based on changepoint detection~\cite{van2020evaluation}. A changepoint is detected through a sliding window average of the 3rd derivative (jerk) of a data stream. The sliding window average has several parameters $p$ which are determined by the user. These include the size of the sliding window, the minimum size of a segment, and the threshold for constructing a new segment. These changepoints are recorded for each data stream, then probabilistically combined into a single set of keypoints for the entire demonstration. To probabilistically combine the changepoints, we first convert the discovered changepoints to probabilistic keypoints $k_\pi$. Each changepoint is probabilistically modeled as a Gaussian $\mathcal{N}(\mu, \sigma)$ where $\mu$ is the location of the changepoint and $\sigma$ is determined by the window size used for changepoint detection. Then, each set of probabilistic keypoints is combined into a single set of probabilistic keypoints for the entire demonstration $\mathcal{K}_\pi$ by finding the product of all the individual sets of keypoints. Finally, these combined probabilistic keypoints are used to find the keypoints of the overall demonstration through sampling, where samples are taken from the probabilistic keypoints until a single set of common keypoints $\mathcal{K}$ is found. This single set of keypoints is then used to segment the entire demonstration into motion primitives.

\subsection{Clustering}

\begin{algorithm}[t]
\DontPrintSemicolon
\small
\label{clustering_alg}
  \KwInput{Data $\vect{\zeta}$, Stretching Constant $\lambda$}
  \KwOutput{Clusters $\vect{\kappa}$}
    $N = 0$\;
    \While{Energy Decreases}
    {  
        $N = N + 1$\;
        $\vect{x}$ = Random-Selection($\vect{\zeta}, N$)\;
        \While{not converged}
        {
            $\kappa_i$ = [] for $i = 1...N$\;
            $A$ = [0]$_{N \times N}$\;
            $C$ = [0]$_{N}$\;
            \For{$i = 1...N$}
            {
                \For{$j = i...N$}
                {
                    $A_{i, i} = A_{i, i} - \lambda$\;
                    $A_{j, j} = A_{j, j} - \lambda$\;
                    $A_{i, j} = A_{i, j} + \lambda$\;
                    $A_{j, i} = A_{j, i} + \lambda$\;
                }
            }
            \For{$\zeta_j \in \vect{\zeta}$}
            {
                $a = \argmin_{i=1...N} ||\zeta_j - x_i||_2$\;
                $\kappa_a \longleftarrow \zeta_j$\;
            }
            \For{$i = 1...N$}
            {
                $A_{i, i} = A_{i, i} + |\kappa_i|$\;
                $C_i = \sum_{\zeta_j \in \kappa_i} \zeta_j $\;
            }
            $\vect{x}$ = $A^{-1} C$\;
        }
        Energy = $U_\mathcal{X} + U_E$ \tcp{Using eq. (1) and (2)}
    }

\caption{Elastic Clustering Algorithm}
\end{algorithm}

\begin{figure*}[ht]
    \centering
    \includegraphics[width=0.98\linewidth]{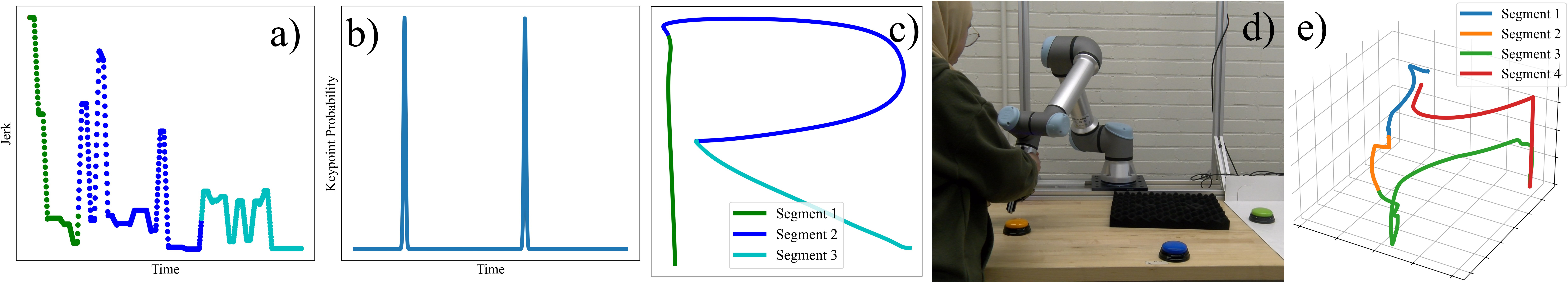}
    
    \caption{\small{Segmenting a 2D handwritten R shape and a real-world demonstration of pressing multiple buttons. a) The jerk profile of the shape, with changepoint segments shown in different colors. b) the probabilities of keypoints in time. c) the segmented shape with different segments shown in different colors. d) The demonstration for pressing multiple buttons on a real-world UR5e robot. e) Multimodal segmentation performed on a real-world demonstration of pressing multiple buttons.}}
    \label{fig:r-segmented}
\end{figure*}

Our next contribution is a novel clustering method, known as elastic clustering\footnote{Clustering module available at: \url{https://github.com/brenhertel/Elastic-Clustering/}}  (shown in Algorithm 1), which is well-suited to clustering robot skills for a library of motion primitives. This clustering algorithm is based on elastic maps~\cite{gorban_zinovyev}, a method for nonlinear dimensionality reduction which has found use for many different applications such as predicting election outcomes~\cite{gorban_visualization2001} and even Learning from Demonstration~\cite{hertel2022ElMap}. Elastic maps find a mean to data by fitting nodes to the representative data. These nodes are connected to the data and other nodes through ``springs.'' By minimizing energies associated with these springs, an elastic map representation of the data is found. Here, we use specific properties of elastic maps to model clusters of data, where each node represents a cluster center.Each node is connected to its cluster through a series of springs, represented by the approximation energy $U_\mathcal{X}$ as
\begin{equation}
     U_\mathcal{X} = \sum_{i = 1}^{N} \sum_{\zeta_j \in \kappa_i} || \zeta_j - x_i ||_2^2, \label{eq:U_X}
\end{equation}
\noindent where $\vect{\zeta} = [\zeta_1, \zeta_2, ..., \zeta_M]$ is the data, $\vect{x} = [x_1, x_2, ..., x_N]$ is the nodes in the elastic map, $\kappa_i$ is the cluster of data for node $x_i$, and $||\cdot||_n$ is the $L^n$-norm. Using only $U_\mathcal{X}$, cluster centers would be weighted averages of the clustered data (as in k-means clustering~\cite{macqueen1967some}). We would be unable to determine the correct number of clusters for a given set of data. Additionally, for robot skills, clusters that are close in the feature space may be undesirable, as they are likely slight modifications of the same skill, and should be clustered together. Therefore, we include another energy in our map, known as the stretching energy $U_E$. The stretching energy pushes cluster centers away from each other as
\begin{equation}
    U_E = -\lambda \sum_{i = 1}^{N} \sum_{j = i}^{N} || x_i - x_j ||_2^2, \label{eq:U_E}
\end{equation}
\noindent where $\lambda$ is the stretching constant. Finally, the optimal map is found by optimizing these two energies as
\begin{equation}
    f(\vect{x}^*) = \underset{\vect{x}}{\text{minimize }}  \sum_{i \in \{ \mathcal{X}, E \}} U_i . \label{eq:opt}
\end{equation}
This minimization can also be performed using an Expectation-Maximization (EM) algorithm~\cite{hertel2022ElMap} as shown in lines 5-21 of Algorithm 1. In this EM formulation, the expectation is performed by clustering the data to the current set of nodes (lines 15-17), then the node positions are optimized according to the approximation and stretching energies determined by the clusters and other nodes (lines 18-21). This optimization is performed using least squares using the inverse of an energy application matrix $A$ with a matrix determined by the clustered nodes $C$.

To cluster robot skills, we iteratively increase the number of clusters $N$ until the minimum energy is found. The data to represent is features of the given motion primitives. For each $N$, we calculate the optimal centers $\vect{x}$ and the energy associated with the optimal map. Once increasing $N$ no longer decreases the optimal map's energy, the optimal number of cluster centers $N$ has been automatically discovered. These clusters are then stored in the library of motion primitives.

\subsection{Selection of Reusable Primitives}

Finally, we propose a method for selecting reusable skills from the library of motion primitives using trajectory editing. A motion primitive library may have many examples of a skill such as pressing a button, but not all pressing examples may be relevant to the current execution environment. Therefore, we use the constraints of the current task (including initial points, final points, via-points, obstacles, etc.) to discover which known skills are appropriate for creating reproductions. To transform skills learned in different environments, we use Laplacian Trajectory Editing (LTE)~\cite{Nierhoff2016LTE}, which warps trajectories to adhere to certain constraints, while maintaining the shape of the original trajectory as closely as feasible. Then, we find a certain number of the best candidate trajectories, determined by the curve length, for use as demonstrations to provide a given LfD representation. As different representations require a different number of demonstrations (i.e., single~\cite{Meirovitch2016JA}, multiple~\cite{Paraschos2013ProMP}, or either~\cite{hertel2022ElMap}), the number of trajectories found is determined by the representation.

\section{Experiments}
\label{sec:exps}

In this paper, we validate the proposed multimodal segmentation and the elastic clustering algorithms individually. We use a variety of simulated and real-world data in these experiments. Unless otherwise specified, real-world data is taken through kinesthetic teaching using a Universal Robots UR5e 6DOF manipulator arm, with an attached Robotiq 2f-85 2-finger gripper.

\subsection{Segmentation}

To validate our segmentation method, we first use a simulated 2D trajectory of the letter R. In this experiment, we do not use multimodal segmentation but rather rely on only a single mode to validate our method. The process of segmentation is shown in Fig.~\ref{fig:r-segmented}a-c. First, changepoint detection is performed on the shape. In this experiment we use a sliding window size of 16, a segment size of 64, and a threshold of 0.16. As seen in Fig.~\ref{fig:r-segmented}a, this creates 2 changepoints within the trajectory. After probabilistically modeling these keypoints (Fig.~\ref{fig:r-segmented}b) and sampling the probability, we find the segments shown in Fig.~\ref{fig:r-segmented}c, which segments the shape at each corner. Visually, this segmentation produces desirable results for this shape.

Next, we perform segmentation on a demonstration taken using a real-world robot. In this demonstration, three buttons are placed in various locations in the robot's workspace. The robot is guided to press each of these buttons in sequence. The demonstration and results of segmenting this demonstration can be seen in Fig.~\ref{fig:r-segmented}d-e. Here, joint data, task-space data, end-effector force data, and gripper position data are all combined to segment the given demonstration. Four segments are found, corresponding to reaching towards each of the three buttons and a final segment for returning to the final position. This shows that multimodal segmentation works on real-world demonstrations, properly segmenting complex motions into motion primitives.

\begin{figure}[ht]
    \centering
    \includegraphics[width=0.98\linewidth]{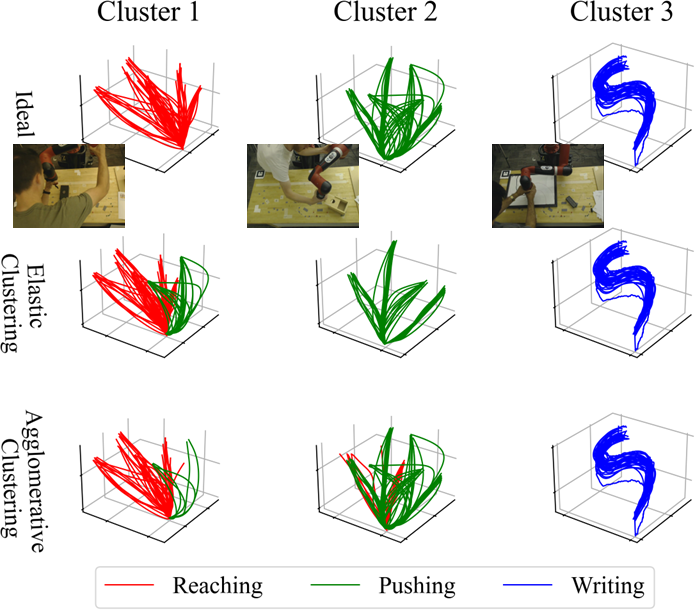}
    \caption{\small{Results of elastic clustering and agglomerative clustering applied to the RAIL Dataset~\cite{rana2020benchmark}. Elastic clustering provides better results as less demonstrations are wrongly clustered.}}
    \label{fig:rail-cluster}
\end{figure}

\subsection{Clustering}

Next, we validate our clustering using real-world demonstrations provided by the RAIL dataset~\cite{rana2020benchmark}. In this dataset, users are asked to perform four different motion primitives using kinesthetic teaching: reaching, pushing, pressing, and writing (images of demonstrations are shown in Fig.~\ref{fig:rail-cluster}). The data provided is already labeled, so we first remove the labels and use all reaching, pushing, and writing demonstrations (150 total). For each demonstration, we first decompose the demonstration into features. The features we use are similarity to a representative demonstration from each class. Once each demonstration is turned into features, it is then passed to the clustering algorithm and processed into clusters. Additionally, we compare elastic clustering against agglomerative clustering~\cite{nielsen2016hierarchical}. Agglomerative clustering is a bottom-up clustering approach, where initially each data point is treated as its own cluster, and then clusters are merged together based on some distance metric. This creates a dendrogram of clusters, which is cut at some level to achieve the desired amount of clusters or distance between clusters. We use the ward linkage criterion and set the number of clusters found to 3. Note that, unlike agglomerative clustering, elastic clustering does not need a parameter for the number of clusters and automatically discovers this using the energy terms of the elastic map. In this experiment, we measure the number of demonstrations in a cluster where that type of demonstration is not the majority, indicating a misclustered demonstration. Our results show that elastic clustering misclusters 20 primitives, or 13.3\% of the dataset. However, agglomerative clustering misclusters 22 demonstrations, or 14.6\% of the dataset. The three clusters found by each algorithm are shown in Fig.~\ref{fig:rail-cluster}. Reaching and pushing motions are very similar and difficult to determine, but elastic clustering is able to differentiate them better than the agglomerative clustering, providing better results overall.

\section{Future Work}

We plan to perform experiments using a real-world Universal Robots UR5e manipulator arm. We will demonstrate several complicated skills using this manipulator arm, including setting the table and making a drink. All demonstrations will be segmented into primitives and clustered. Sparse labels are provided to the clusters (i.e., some segments are labeled as ``pressing,'' ``reaching,'' ``placing,'' etc.). Then, a reproduction of unloading a dishwasher will be asked for. We will use a planner such as Planning Domain Definition Language (PDDL)~\cite{ghallab1998pddl} for high-level task planning. The robot will be able to use the skills learned from the demonstrated tasks to be able to complete a complicated skill that has not been demonstrated.

\section*{Acknowledgements}

This work was supported in part by Amazon, the U.S. Office of Naval Research (N00014-21-1-2582 and N00014-23-1-2744) and National Science Foundation (FRR-
2237463).

\typeout{}
\bibliographystyle{IEEEtran}
\bibliography{references}

\end{document}